\documentclass[11pt,a4paper]{article}
\usepackage[hyperref]{acl2021}
\usepackage{times}
\usepackage{latexsym}

\usepackage{graphicx}
\usepackage{soul}
\usepackage{url}
\usepackage{arydshln}
\usepackage[utf8]{inputenc}
\usepackage{caption}
\usepackage{amsmath}
\usepackage{amsthm}
\usepackage{booktabs}
\usepackage{algorithm}
\usepackage{algorithmic}
\usepackage{multirow}

\usepackage{microtype}

\aclfinalcopy 

\title{Defending Pre-trained Language Models\\from Adversarial Word Substitution Without Performance Sacrifice}

 \author{Rongzhou Bao, Jiayi Wang, Hai Zhao\footnotemark[1]\\
Department of Computer Science and Engineering, Shanghai Jiao Tong University\\
\textsuperscript{2} Key Laboratory of Shanghai Education Commission for Intelligent Interaction\\
and Cognitive Engineering, Shanghai Jiao Tong University, Shanghai, China\\
MoE Key Lab of Artificial Intelligence, AI Institute,\\ Shanghai Jiao Tong University, Shanghai, China\\
\texttt{rongzhou.bao@outlook.com}\\
\texttt{\{wangjiayi\_102\_23\}@sjtu.edu.cn,zhaohai@cs.sjtu.edu.cn}\\
}

\date{}

\begin{document}
\maketitle

\renewcommand{\thefootnote}{\fnsymbol{footnote}}
\footnotetext[1]{ Corresponding author. This paper was partially sup- ported by National Key Research and Development Pro- gram of China (No. 2017YFB0304100), Key Projects of National Natural Science Foundation of China (U1836222 and 61733011). This work was supported by Huawei Noah's Ask Lab} 
\begin{abstract}
Pre-trained contextualized language models (PrLMs) have led to strong performance gains in downstream natural language understanding tasks. However, PrLMs can still be easily fooled by adversarial word substitution, which is one of the most challenging textual adversarial attack methods. Existing defence approaches suffer from notable performance loss and complexities. Thus, this paper presents a compact and performance-preserved framework, \textbf{A}nomaly \textbf{D}etection with \textbf{F}requency-\textbf{A}ware \textbf{R}andomization (ADFAR). In detail, we design an auxiliary anomaly detection classifier and adopt a multi-task learning procedure, by which PrLMs are able to distinguish adversarial input samples. Then, in order to defend adversarial word substitution, a frequency-aware randomization process is applied to those recognized adversarial input samples. Empirical results show that ADFAR significantly outperforms those newly proposed defense methods over various tasks with much higher inference speed. Remarkably, ADFAR does not impair the overall performance of PrLMs. The code is available at \href{https://github.com/LilyNLP/ADFAR}{https://github.com/LilyNLP/ADFAR}.

\end{abstract}

\section{Introduction}
Deep neural networks (DNNs) have achieved remarkable success in various areas. However, previous works show that DNNs are vulnerable to adversarial samples \cite{goodfellow2015explaining,kurakin2017adversarial,wang2021robust}, which are inputs with small, intentional modifications that cause the model to make false predictions. Pre-trained language models (PrLMs) \cite{devlin:bert,liu2019roberta,Clark2020ELECTRA:,zhang2020semanticsaware,zhang2019sgnet} are widely adopted as an essential component for various NLP systems. However, as DNN-based models, PrLMs can still be easily fooled by textual adversarial samples \cite{Wallace2019Triggers,jin2019bert,nie-etal-2020-adversarial,zang-etal-2020-word}. Such vulnerability of PrLMs keeps raising potential security concerns, therefore researches on defense techniques to help PrLMs against textual adversarial samples are imperatively needed.

Different kinds of textual attack methods have been proposed, ranging from character-level word misspelling \cite{Black-Box}, word-level substitution \cite{Alzantot,ebrahimi-etal-2018-adversarial,ren-etal-2019-generating,jin2019bert,zang-etal-2020-word,li-etal-2020-bert-attack,garg-ramakrishnan-2020-bae}, phrase-level insertion and removal \cite{Liang_2018}, to sentence-level paraphrasing \cite{sears:acl18,Iyyer}. Thanks to the discrete nature of natural language, attack approaches that result in illegal or unnatural sentences can be easily detected and restored by spelling correction and grammar error correction \cite{10.5555/1699648.1699670,SakaguchiPD17,pruthi-etal-2019-combating}. However, attack approaches based on adversarial word substitution can produce high-quality and efficient adversarial samples which are still hard to be detected by existing methods. Thus, the adversarial word substitution keeps posing a larger and more profound challenge for the robustness of PrLMs. Therefore, this paper is devoted to overcome the challenge posed by adversarial word substitution.

Several approaches are already proposed to mitigate issues posed by adversarial word substitution \cite{ZhouJCW19,JiaRGL19,HuangSWDYGDK19,pmlr-v97-cohen19c,ye-etal-2020-safer,si2021better}. Although these defense methods manage to alleviate the negative impact of adversarial word substitution, they sometimes reduce the prediction accuracy for non-adversarial samples to a notable extent. Given the uncertainty of the existence of attack in real application, it is impractical to sacrifice the original prediction accuracy for the purpose of defense. Moreover, previous defense methods either have strong limitations over the attack space to certify the robustness, or require enormous computation resources during training and inference. Thus, it is imperatively important to find an efficient performance-preserved defense method. 

For such purpose, we present a compact and performance-preserved framework, \textbf{A}nomaly \textbf{D}etection with \textbf{F}requency-\textbf{A}ware \textbf{R}andomization (ADFAR), to help PrLMs defend against adversarial word substitution without performance sacrifice. \citet{xie2018mitigating} show that introducing randomization at inference can effectively defend adversarial attacks. Moreover, \cite{mozes2020frequencyguided} indicate that the usual case for adversarial samples is replacing words with their less frequent synonyms, while PrLMs are more robust to frequent words. Therefore, we propose a frequency-aware randomization process to help PrLMs defend against adversarial word substitution. 

However, simply applying a randomization process to all input sentences would reduce the prediction accuracy for non-adversarial samples. In order to preserve the overall performance, we add an auxiliary anomaly detector on top of PrLMs and adopt a multi-task learning procedure, by which PrLMs are able to determine whether each input sentence is adversarial or not, and not introduce extra model. Then, only those adversarial input sentences will undergo the randomization procedure, while the prediction process for non-adversarial input sentences remains the same. 

Empirical results show that as a more efficient method, ADFAR significantly outperforms previous defense methods \cite{ye-etal-2020-safer,ZhouJCW19} over various tasks, and preserves the prediction accuracy for non-adversarial sentences. Comprehensive ablation studies and analysis further prove the efficiency of our proposed method, and indicate that the adversarial samples generated by current heuristic word substitution strategy can be easily detected by the proposed auxiliary anomaly detector. 


\begin{figure*}[htb!]
	\centering
	\includegraphics[width=1.0\textwidth]{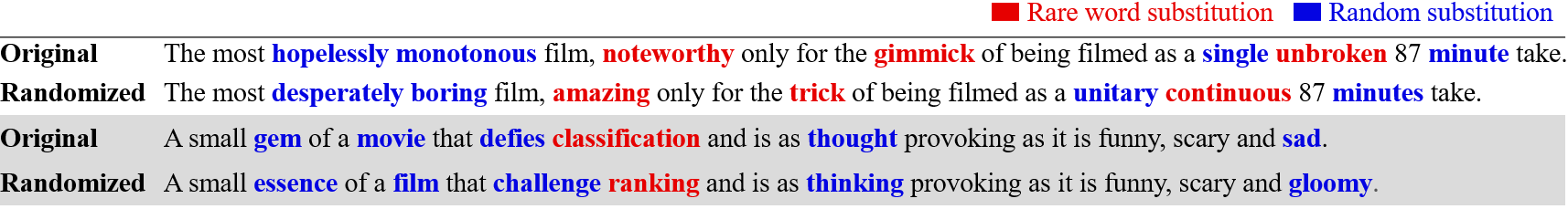}
	\caption{\label{fig:examples}Frequency-aware randomization examples.}
\end{figure*}

\section{Related Work}
\subsection{Adversarial Word Substitution}
Adversarial word substitution (AWS) is one of the most efficient approaches to attack advanced neural models like PrLMs. In AWS, an attacker deliberately replaces certain words by their synonyms to mislead the prediction of the target model. At the same time, a high-quality adversarial sample should maintain grammatical correctness and semantic consistency. In order to craft efficient and high-quality adversarial samples, an attacker should first determine the vulnerable tokens to be perturbed, and then choose suitable synonyms to replace them. 

Current AWS models \cite{Alzantot,ebrahimi-etal-2018-adversarial,ren-etal-2019-generating,jin2019bert,li-etal-2020-bert-attack,garg-ramakrishnan-2020-bae} adopt heuristic algorithms to locate vulnerable tokens in sentences. To illustrate, for a given sample and a target model, the attacker iteratively masks the tokens and checks the output of the model. The tokens which have significant influence on the final output logits are regarded as vulnerable.

Previous works leverage word embeddings such as GloVe \cite{pennington2014glove} and counter-fitted vectors \cite{2016counterfitting} to search the suitable synonym set of a given token. \citet{li-etal-2020-bert-attack,garg-ramakrishnan-2020-bae} uses BERT \cite{devlin:bert} to generate perturbation for better semantic consistency and language fluency.

\subsection{Defense against AWS}
For general attack approaches, adversarial training \cite{goodfellow2015explaining,jiang-etal-2020-smart} is widely adopted to mitigate adversarial effect, but \cite{Alzantot,jin2019bert} shows that this method is still vulnerable to AWS. This is because AWS models leverage dynamic algorithms to attack the target model, while adversarial training only involves a static training set.

Methods proposed by \citet{JiaRGL19,HuangSWDYGDK19} are proved effective for defence against AWS, but they still have several limitations. In these methods, Interval Bound Propagation (IBP) \cite{Krishnamurthy}, an approach to consider the worst-case perturbation theoretically, is leveraged to certify the robustness of models. However, IBP-based methods can only achieve the certified robustness under a strong limitation over the attack space. Furthermore, they are difficult to adapt to PrLMs for their strong reliance on the assumption of model architecture.

Two effective and actionable methods (DISP \cite{ZhouJCW19} and SAFER \citet{ye-etal-2020-safer}) are proposed to overcome the challenge posed by AWS, and therefore adopted as the baselines for this paper. DISP \cite{ZhouJCW19} is a framework based on perturbation discrimination to block adversarial attack. In detail, when facing adversarial inputs, DISP leverages two auxiliary PrLMs: one to detect perturbed tokens in the sentence, and another to restore the abnormal tokens to original ones. Inspired by randomized smoothing \cite{pmlr-v97-cohen19c}, \citet{ye-etal-2020-safer} proposes SAFER, a novel framework that guarantees the robustness by smoothing the classifier with synonym word substitution. To illustrate, based on random word substitution, SAFER smooths the classifier by averaging its outputs of a set of randomly perturbed inputs. SAFER outperforms IBP-based approaches and can be easily applied to PrLMs.

\subsection{Randomization}
In recent years, randomization has been used as a defense measure for deep learning in computer vision \cite{xie2018mitigating}. Nevertheless, direct extensions of these measures to defend against textual adversarial samples are not achievable, since the text inputs are discrete rather than continuous. \citet{ye-etal-2020-safer} indicates the possibility of extending the application of the randomization approach to NLP by randomly replacing the words in sentences with their synonyms.

\section{Method}
\subsection{Frequency-aware Randomization}

Since heuristic attack methods attack a model by substituting each word iteratively until it successfully alters the model's output, it is normally difficult for static strategies to defense such kind of dynamic process. Rather, dynamic strategies, such as randomization, can better cope with the problem. It is also observed that replacing words with their more frequent alternatives can better mitigate the adversarial effect and preserve the original performance. Therefore, a frequency-aware randomization strategy is designed to perplex AWS strategy. 

Figure \ref{fig:examples} shows several examples of the frequency-aware randomization. The proposed approach for the frequency-aware randomization is shown in Algorithm \ref{alg1}, and consists of three steps. Firstly, rare words with lower frequencies and a number of random words are selected as substitution candidates. Secondly, we choose synonyms with the closest meanings and the highest frequencies to form a synonym set for each candidate word. Thirdly, each candidate word is replaced with a random synonym within its own synonym set. To quantify the semantic similarity between two words, we represent words with embeddings from \cite{2016counterfitting}, which is specially designed for synonyms identification. The semantic similarity of two words are evaluated by cosine similarity of their embeddings. To determine the frequency of a word, we use a frequency dictionary provided by FrequencyWords Repository\footnotemark[1].\footnotetext[1]{https://github.com/hermitdave/FrequencyWords}

\begin{algorithm}
	\renewcommand{\algorithmicrequire}{\textbf{Input:}}
	\renewcommand{\algorithmicensure}{\textbf{Output:}}
	\caption{Frequency-aware Randomization}
	\label{alg1}
	\begin{algorithmic}[1]
	    \REQUIRE Sentence $X=\{w_1, w_2, ..., w_n\}$, word embeddings $Emb$ over the vocabulary $Vocab$
	    \ENSURE Randomized sentence $X_{rand}$
		\STATE Initialization: $X_{rand} \leftarrow X$
		\STATE Create a set $W_{rare}$ of all rare words with frequencies less than $f_{thres}$, denote $n_{rare} = |W_{rare}|$.
		\STATE Create a set $W_{rand}$ by randomly selecting $n*r-n_{rare}$ words $w_j \notin W_{rare}$, where $r$ is the pre-defined ratio of substitution.
		\STATE Create the substitution candidates set, $W_{sub} \leftarrow W_{rare} + W_{rand}$, and $|W_{sub}| = n*r$.
		\STATE Filter out the stop words in $W_{sub}$.
		\FOR{each word $w_i$ in $W_{sub}$}
		\STATE Create a set $S$ by extracting the top $n_s$ synonyms using $CosSim(Emb_{w_i}, Emb_{w_{word}})$ for each word in $Vocab$.
		\STATE Create a set $S_{freq}$ by selecting the top $n_f$ frequent synonyms from  $S$.
		\STATE Randomly choose one word $w_s$ from $S$.
		\STATE $X_{rand} \leftarrow$ Replace $w_i$ with $w_s$ in $X_{rand}$.
		\ENDFOR
	\end{algorithmic}  
\end{algorithm}


\subsection{Anomaly Detection}
Applying the frequency-aware randomization process to every input can still reduce the prediction accuracy for normal samples. In order to overcome this issue, we add an auxiliary anomaly detection head to PrLMs and adopt a multi-task learning procedure, by which PrLMs are able to classify the input text and distinguish the adversarial samples at the same time, and not introduce extra model. In inference, the frequency-aware randomization only applied to the samples that are detected as adversarial. In this way, the reduction of accuracy is largely avoided, since non-adversarial samples are not affected.

\citet{ZhouJCW19} also elaborates the idea of perturbation discrimination to block attack. However, their method detects anomaly on token-level and requires two resource-consuming PrLMs for detection and correction, while ours detects anomaly on sentence-level and requires no extra models. Compared to \citet{ZhouJCW19}, our method is two times faster in inference speed and can achieve better accuracy for sentence-level anomaly detection.

\subsection{Framework}
In this section, we elaborate the framework of ADFAR in both training and inference. 
\subsubsection{Training}

\begin{figure}
	\centering
	\includegraphics[width=0.35\textwidth]{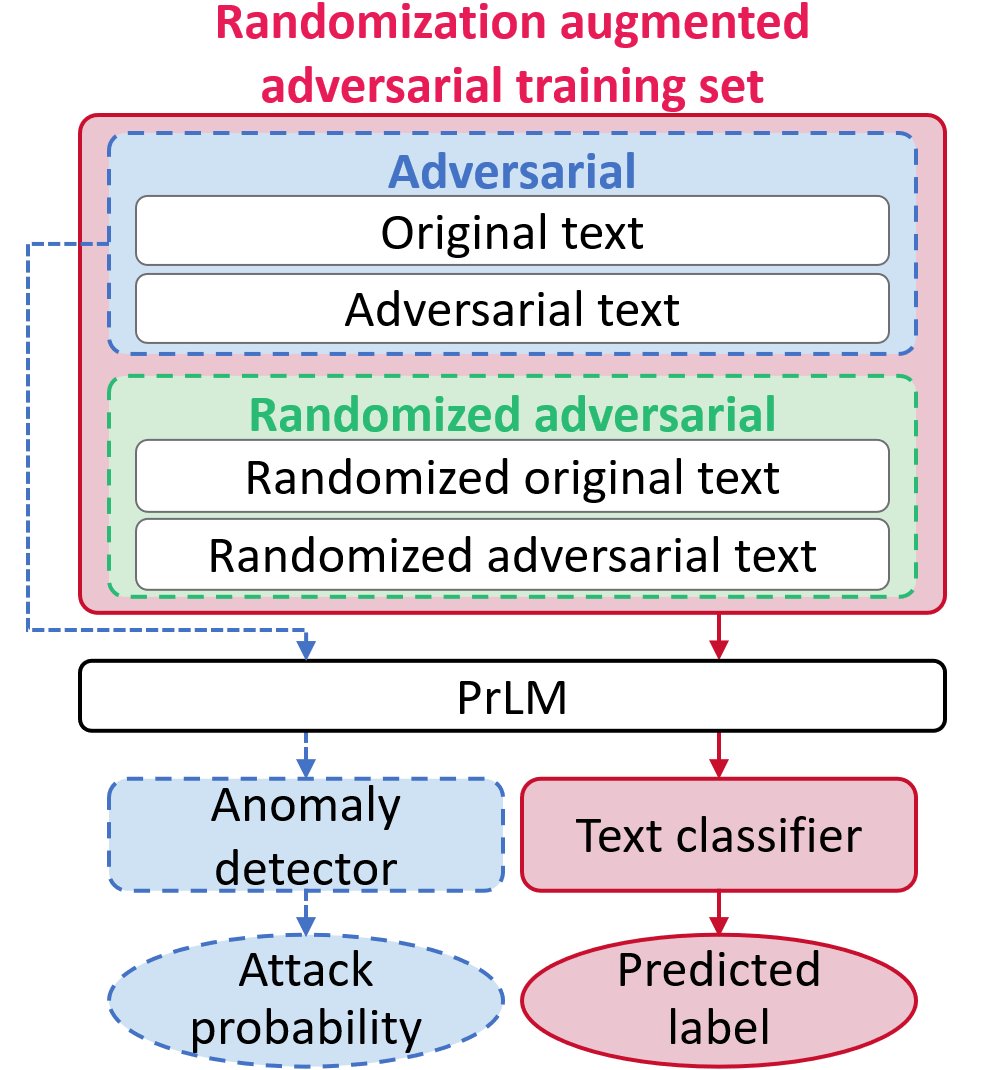}
	\caption{\label{fig:training stage}Framework of ADFAR in training.}
\end{figure}

Figure \ref{fig:training stage} shows the framework of ADFAR in  training. We extend the baseline PrLMs by three major modifications: 1) the construction of training data, 2) the auxiliary anomaly detector and 3) the training objective, which will be introduced in this section.

\paragraph{Construction of Training Data}
As shown in Figure \ref{fig:training stage}, we combine the idea of both adversarial training and data augmentation \cite{wei-zou-2019-eda} to construct our randomization augmented adversarial training data. Firstly, we use a heuristic AWS model (e.g. TextFooler) to generate adversarial samples based on the original training set. Following the common practice of adversarial training, we then combine the adversarial samples with the original ones to form an adversarial training set. Secondly, in order to let PrLMs better cope with randomized samples in inference, we apply the frequency-aware randomization on the adversarial training set to generate a randomized adversarial training set. Lastly, the adversarial training set and the randomized adversarial training set are combined to form a randomization augmented adversarial training set.

\paragraph{Auxiliary Anomaly Detector}
In addition to the original text classifier, we add an auxiliary anomaly detector to the PrLMs to distinguish adversarial samples. For an input sentence, the PrLMs captures the contextual information for each token by self-attention and generates a sequence of contextualized embeddings $\{h_0, \dots, h_m\}$. For text classification task, $h_0 \in R^H$ is used as the aggregate sequence representation. The original text classifier leverages $h_0$ to predict the probability that $X$ is labeled as class $\hat{y_c}$ by a logistic regression with softmax:
\begin{equation}\nonumber
\begin{aligned}
y_c &= Prob(\hat{y_c}|x), \\
&= {\rm softmax}(W_c(dropout(h_0))+b_c),  
\end{aligned}
\end{equation}

For the anomaly detector, the probability that $X$ is labeled as class $\hat{y_d}$ (if $X$ is attacked, $\hat{y_d} = 1$; if $X$ is normal, $\hat{y_d} = 0$) is predicted by a logistic regression with softmax:
\begin{equation}\nonumber
\begin{aligned}
y_d &= Prob(\hat{y_d}|x), \\
&= {\rm softmax}(W_d(dropout(h_0))+b_d),   
\end{aligned}
\end{equation}
As shown in Figure \ref{fig:training stage}, the original text classifier is trained on the randomization augmented adversarial training set, whereas the anomaly detector is only trained on the adversarial training set.

\paragraph{Training Objective}
We adopt a multi-task learning framework, by which PrLM is trained to classify the input text and distinguish the adversarial samples at the same time. We design two parallel training objectives in the form of minimizing cross-entropy loss: $loss_c$ for text classification and $loss_d$ for anomaly detection. The total loss function is defined as their sum:
\begin{equation}\nonumber
\begin{split}
loss_c &= -[y_c*{\rm log}(\hat{y_c})+(1-y_c)*{\rm log}(1-\hat{y_c})] \\
loss_d &= -[y_d*{\rm log}(\hat{y_d})+(1-y_d)*{\rm log}(1-\hat{y_d})] \\
Loss &= loss_c + loss_d
\end{split}
\end{equation}

\subsubsection{Inference}

\begin{figure}
	\centering
	\includegraphics[width=0.48\textwidth]{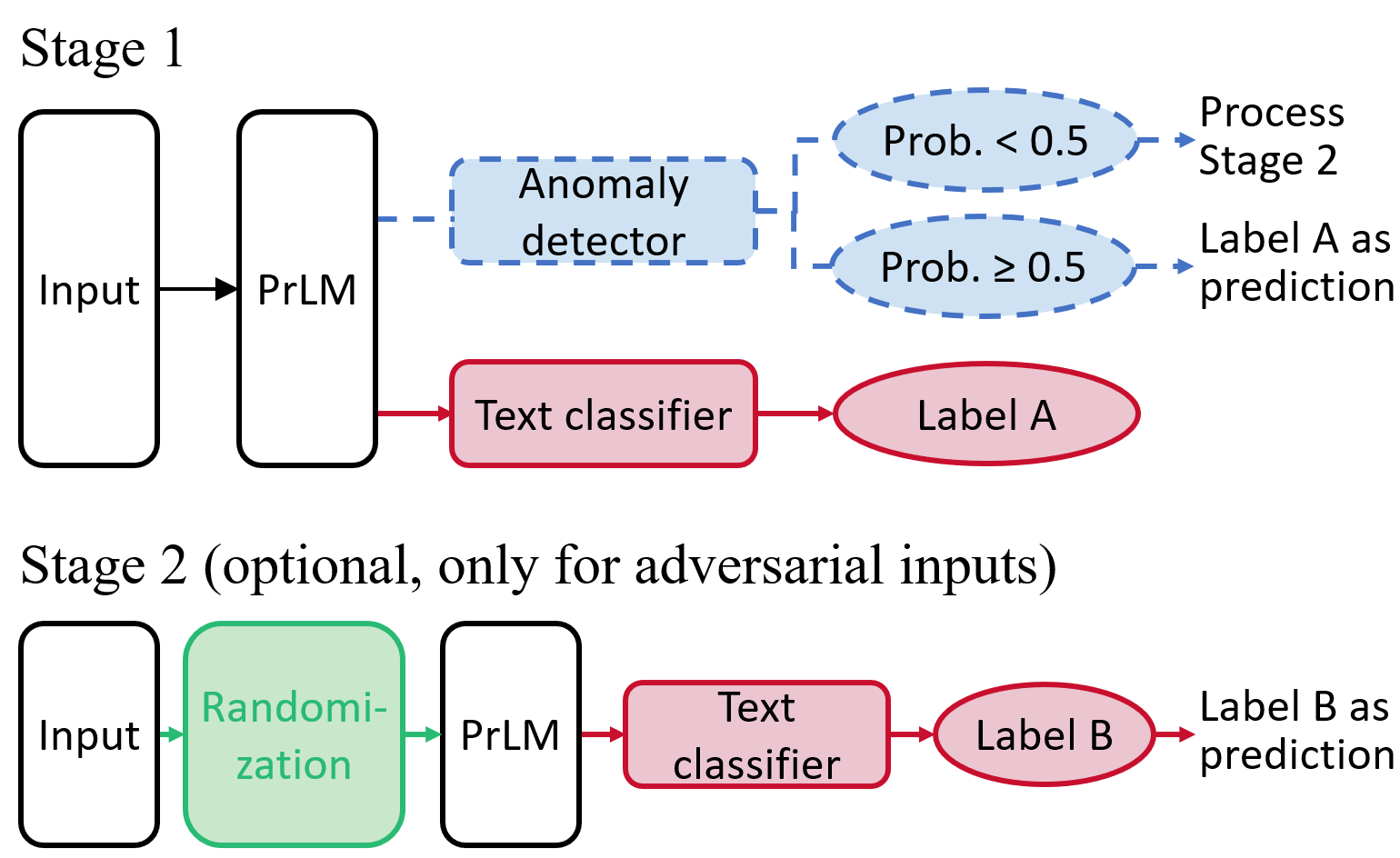}
	\caption{\label{fig:inference stage}Framework of ADFAR in inference.}
\end{figure}

Figure \ref{fig:inference stage} shows the framework of ADFAR in inference. Firstly, the anomaly detector predicts whether an input sample is adversarial. If the input sample is determined as non-adversarial, the output of the text classifier (Label A) is directly used as its final prediction. If the input sample is determined as adversarial, the frequency-aware randomization process is applied to the original input sample. Then, the randomized sample is sent to the PrLM again, and the second output of the text classifier (Label B) is used as its final prediction.

\section{Experimental Implementation}
\subsection{Tasks and Datasets}
Experiments are conducted on two major NLP tasks: text classification and natural language inference. The dataset statistics are displayed in Table \ref{tab:compairison}. We evaluate the performance of models on the non-adversarial test samples as the original accuracy. Then we measure the after-attack accuracy of models when facing AWS. By comparing these two accuracy scores, we can evaluate how robust the model is.

\begin{table}[htb!]
    \centering
    \small
    \setlength{\tabcolsep}{3pt}
    {
        \begin{tabular}{clccc}
        \toprule
        Task & Dataset & Train & Test & Avg Len\\
        \midrule
        \multirow{3}{*}{Classification} & MR & 9K & 1K & 20 \\
        &SST2 & 67K & 1.8K & 20 \\
        &IMDB & 25K & 25K & 215 \\
        \midrule
        Entailment & MNLI & 433K & 10K & 11 \\
        \bottomrule
        \end{tabular}
    }
\caption{\label{tab:compairison} Dataset statistics.}
\label{tab:booktabs}
\end{table}

\paragraph{Text Classification}
We use three text classification datasets with average text lengths from 20 to 215 words, ranging from phrase-level to document-level tasks. \textbf{SST2} \cite{socher-etal-2013-recursive}: phrase-level binary sentiment classification using fine-grained sentiment labels on movie reviews. \textbf{MR} \cite{pang-lee-2005-seeing}: sentence-level binary sentiment classification on movie reviews. We take 90$\%$ of the data as training set and 10$\%$ of the data as test set as \cite{jin2019bert}. \textbf{IMDB} \cite{maas-etal-2011-learning}: document-level binary sentiment classification on movie reviews.

\paragraph{Natural Language Inference}  NLI aims at determining the relationship between  a  pair  of  sentences based  on  semantic meanings. We use Multi-Genre Natural Language Inference (\textbf{MNLI}) \cite{nangia2017repeval}, a widely adopted NLI benchmark with coverage of transcribed speech, popular fiction, and government reports.

\begin{table*}[htb!]
	\centering
	\small
	\setlength{\tabcolsep}{3pt}
	{
        \begin{tabular}{l|cc|cc|cc|cc}
        \toprule
        
        \multirow{2}{*}{Model} &
        \multicolumn{2}{c|}{MR} &
        \multicolumn{2}{c|}{SST2} &
        \multicolumn{2}{c|}{IMDB} &
        \multicolumn{2}{c}{MNLI} \\
        &
        Orig. Acc. &
        Adv. Acc. &
        Orig. Acc. &
        Adv. Acc. &
        Orig. Acc. &
        Adv. Acc. &
        Orig. Acc. &
        Adv. Acc. \\
        \hline
        BERT                 & 86.2 & 16.9 & \textbf{93.1} & 39.8 & 92.4  &12.4  & \textbf{84.0}  & 11.3  \\
        BERT + Adv Training &   85.6   &   34.6   & 92.6 & 48.8 & 92.2 & 34.2 & 82.3 & 33.4 \\
        BERT + DISP                 &   82.0   &   42.2   & 91.6     & 70.4     & 91.7 & 82.0 & 76.3 &  35.1\\
        BERT + SAFER                &   79.0   &  55.4    &  91.3    & 75.6     &  91.3 & 88.1  & 82.1  & 54.7 \\
        \hdashline
        BERT + ADFAR                 & \textbf{86.6} & \textbf{66.0} & 92.4 & \textbf{75.6} & \textbf{92.8} & \textbf{89.2} & 82.6 & \textbf{67.8} \\
        \bottomrule
        
        \end{tabular}
    }
    \caption{\label{tab:main_result} The performance of ADFAR and other defense frameworks using BERT$_\text{BASE}$ as PrLM and TextFooler as attack model. Orig. Acc. is the prediction accuracy of normal samples and Adv. Acc. is the after-attack accuracy of models when facing AWS. The results are based on the average of five runs. }
\end{table*}

\subsection{Attack Model and Baselines}
We use TextFooler\footnotemark[2]\cite{jin2019bert} as the major attack model for AWS. Moreover, we implement \cite{ren-etal-2019-generating} and GENETIC \cite{Alzantot} based on the TextAttack \cite{morris-etal-2020-textattack} code base to further verify the efficiency of our proposed method.
\footnotetext[2]{https://github.com/jind11/TextFooler}

We compare ADFAR with DISP \cite{ZhouJCW19} and SAFER \cite{ye-etal-2020-safer}. The implementation of DISP is based on the repository\footnotetext[3]{https://github.com/joey1993/bert-defender} offered by \citet{ZhouJCW19}. For SAFER, we also leverage the code\footnotetext[4]{https://github.com/lushleaf/Structure-free-certified-NLP} proposed by \citet{ye-etal-2020-safer}. Necessary modifications are made to evaluate these methods' performance under heuristic attack models.

\subsection{Experimental Setup}
The implementation of PrLMs is based on PyTorch\footnotemark[3]. We leverage, BERT$_\text{BASE}$  \cite{devlin:bert}, RoBERTa$_\text{BASE}$ \cite{liu2019roberta} and ELECTRA$_\text{BASE}$ \cite{Clark2020ELECTRA:} as baseline PrLMs. We use AdamW \cite{DBLP:journals/corr/abs-1711-05101} as our optimizer with a learning rate of 3e-5 and a batch size of 16. The number of epochs is set to 5. 
\footnotetext[5]{https://github.com/huggingface} 

For the frequency-aware randomization process, we set $f_{thres} = 200$, $n_s = 20$ and $n_f = 10$. In the adopted frequency dictionary, 5.5k out of 50k words have a frequency lower than $f_{thres} = 200$ and therefore regarded as rare words. $r$ is set to different values for training ($25\%$) and inference ($30\%$) due to different aims. In training, to avoid introducing excessive noise and reduce the prediction accuracy for non-adversarial samples, $r$ is set to be relatively low. On the contrary, in inference, our aim is to perplex the heuristic attack mechanism. The more randomization we add, the more perplexities the attack mechanism receives, therefore we set a relatively higher value for $r$. More details on the choice of these hyperparameters will be discussed in the analysis section. 

\section{Experimental Results}

\subsection{Main results}

Following \cite{jin2019bert}, we leverage BERT$_\text{BASE}$ \cite{devlin:bert} as baseline PrLM and TextFooler as attack model. Table \ref{tab:main_result} shows the performance of ADFAR and other defense frameworks. Since randomization may lead to a variance of the results, we report the results based on the average of five runs. Experimental results indicate that ADFAR can effectively help PrLM against AWS. Compared with DISP \cite{ZhouJCW19} and SAFER \cite{ye-etal-2020-safer}, ADFAR achieves the best performance for adversarial samples. Meanwhile, ADFAR does not hurt the performance for non-adversarial samples in general. On tasks such as MR and IMDB, ADFAR can even enhance the baseline PrLM. 

DISP leverages two extra PrLMs to discriminate and recover the perturbed tokens, which introduce extra complexities. SAFER makes the prediction of an input sentence by averaging the prediction results of its perturbed alternatives, which multiply the inference time. As shown in Table \ref{tab:methods}, compared with previous methods, ADFAR achieves a significantly higher inference speed.

\begin{table}[htb!]
    \centering
    \small
    \setlength{\tabcolsep}{6pt}
    {
        \begin{tabular}{lcc}
        \toprule
        Model & Parameters & Inference Time \\
        \midrule
        BERT$_\text{BASE}$ & 110M & 15.7ms (100\%) \\
        BERT$_\text{BASE}$ + DISP & 330M & 38.9ms (247\%) \\
        BERT$_\text{BASE}$ + SAFER & 110M &  27.6ms (176\%)\\
        \hdashline
        BERT$_\text{BASE}$ + ADFAR & 110M & 18.1ms (115\%) \\
        \bottomrule
        \end{tabular}
    }
\caption{\label{tab:methods} Parameters and Inference Time statistics. The inference time indicate the average inference time for one sample in MR dataset using one NVIDIA RTX3090.}
\label{tab:booktabs}
\end{table}

\subsection{Results with Different Attack Strategy}

Since ADFAR leverages the adversarial samples generated by TextFooler \cite{jin2019bert} in training, it is important to see whether ADFAR also performs well when facing adversarial samples generated by other AWS models. We leverage PWWS \cite{ren-etal-2019-generating} and GENETIC \cite{Alzantot} to further study the performance of ADFAR. 

\begin{table}[htb!]
	\centering
	\small
	\setlength{\tabcolsep}{6pt}
	{
        \begin{tabular}{l|cc|cc}
        \toprule
        \multirow{2}{*}{Attack} &
        \multicolumn{2}{c|}{MR} &
        \multicolumn{2}{c}{SST2} \\ 
         &
        BERT &
        +ADFAR &
        BERT &
       +ADFAR\\
        
        \hline
        Attack-Free           &    86.2   & 86.6 & 93.1 & 92.3 \\
        PWWS  &  34.2    &   74.2  & 54.3&  80.5  \\
        Genetic  & 21.3  &  70.4 &  38.7  & 72.2 \\
        TextFooler &  16.9 & 66.0  &  39.8 &  73.8   \\
        \bottomrule
        
        \end{tabular}
    }
    \caption{\label{tab:different} Performance of BERT and BERT with ADFAR when facing various AWS models. The results are based on the average of five runs.}
\end{table}

As shown is Table \ref{tab:different}, the performance of ADFAR is not affected by different AWS models, which further proves the efficacy of our method. 

\subsection{Results with Other PrLMs}
Table \ref{tab:PrLMs} shows the performance of ADFAR leveraging RoBERTa$_\text{BASE}$ \cite{liu2019roberta} and ELECTRA$_\text{BASE}$ \cite{Clark2020ELECTRA:} as PrLMs. In order to enhance the robustness and performance of the PrLM, RoBERTa extends BERT with a larger corpus and using more efficient parameters, while ELECTRA applies a GAN-style architecture for pre-training. Empirical results indicate that ADFAR can further improve the robustness of RoBERTa and ELECTRA while preserving their original performance.

\begin{table}[htb!]
	\centering
	\small
	\setlength{\tabcolsep}{3pt}
	{
        \begin{tabular}{l|cc|cc}
        \toprule
        \multirow{2}{*}{PrLM} &
        \multicolumn{2}{c|}{MR} &
        \multicolumn{2}{c}{SST2} \\ 
         &
        Orig. Acc. &
        Adv. Acc. &
        Orig. Acc. &
       Adv. Acc. \\
        
        \hline
        BERT           &    86.2   & 16.9 & 93.1 & 39.8 \\
        \hspace{0.2cm}+ADFAR           &    86.6   & 66.0 & 92.3 & 73.8 \\
        \hdashline
        RoBERTa  &  88.3   &  30.4  &  93.4 & 37.4  \\
        \hspace{0.2cm}+ADFAR    &  87.2   &  71.0  & 93.2 & 77.6 \\
        \hdashline
        ELECTRA  & 90.1 &  33.6 & 94.2 & 40.4 \\
        \hspace{0.2cm}+ADFAR    & 90.4 &  71.2 & 95.0 & 83.0 \\
        
        \bottomrule
        
        \end{tabular}
    }
    \caption{\label{tab:PrLMs} Results based with various PrLMs. }
\end{table}

\section{Analysis}

\subsection{Ablation Study}
ADFAR leverages three techniques to help PrLMs defend against adversarial samples: adversarial training, frequency-aware randomization and anomaly detection. To evaluate the contributions of these techniques in ADFAR, we perform ablation studies on MR and SST2 using BERT$_\text{BASE}$ as our PrLMs, and TextFooler as the attack model. As shown in Table \ref{tab:ablation}, the frequency-aware randomization is the key factor which helps PrLM defense against adversarial samples, while anomaly detection plays an important role in preserving PrLM's prediction accuracy for non-adversarial samples. 
\begin{table}[htb!]
	\centering
	\small
	\setlength{\tabcolsep}{2pt}
	{
        \begin{tabular}{l|cc|cc}
        \toprule
        
        \multirow{2}{*}{Model} &
        \multicolumn{2}{c|}{MR} &
        \multicolumn{2}{c}{SST2} \\
        &
        Orig. Acc. &
        Adv. Acc. &
        Orig. Acc. &
        Adv. Acc. \\
        \hline
        BERT            & 86.2 & 16.9 & 93.1 & 39.8 \\
            \hspace{0.2cm}+ Adv  & 85.6 & 34.6 & 92.6 & 48.8  \\
            \hspace{0.2cm}+ FR            & 85.0 & 72.8 & 90.6 & 82.6 \\
            \hspace{0.2cm}+ AD            & 86.6 & 66.0 & 92.3 & 73.8 \\
        \bottomrule
        
        \end{tabular}
    }
    \caption{\label{tab:ablation} Ablation study on MR and SST2 using BERT$_\text{BASE}$ as PrLM, and TextFooler as attack model. Adv represents adversarial training, FR indicates frequency-aware randomization and AD means anomaly detection. The results are based on the average of five runs.}
\end{table}

\subsection{Anomaly Detection}
In this section, we compare the anomaly detection capability between ADFAR and DISP \cite{ZhouJCW19}. ADFAR leverages an auxiliary anomaly detector, which share a same PrLM with the original text classifier, to discriminate adversarial samples. DISP uses an discriminator based on an extra PrLMs to identify the perturbed adversarial inputs, but on token level. For DISP, in order to detect anomaly on sentence level, input sentences with one or more than one adversarial tokens identified by DISP are regarded as adversarial samples. We respectively sample 500 normal and adversarial samples from the test set of MR and SST to evaluate the performance of ADFAR and DISP for anomaly detection.

Table \ref{tab:abnormal} shows the performance of ADFAR and DISP for anomaly detection. Empirical results show that ADFAR can predict more precisely, since it achieves a significantly higher $F_1$ score than DISP. Moreover, ADFAR has a simpler framework, as its anomaly detector shares the same PrLM with the classifier, while DISP requires an extra PrLM. The results also indicate that the current heuristic AWS strategy is vulnerable to our anomaly detector, which disproves the claimed undetectable feature of this very adversarial strategy. 

\begin{table}[htb!]
	\centering
	\small
	\setlength{\tabcolsep}{2pt}
	{
        \begin{tabular}{c|cll|cll}
        \toprule
        
        \multirow{2}{*}{Method} &
        \multicolumn{3}{c|}{MR} &
        \multicolumn{3}{c}{SST2} \\
        &
        Precision &
        Recall &
        $F_1$ &
        Precision &
        Recall&
        $F_1$\\
        \hline
        DISP  &  68.0 & 92.0 & 73.2 & 59.5 & 94.2 & 72.9 \\
        ADFAR & 90.1 & 84.0 & 86.9 & 88.0 & 90.0 & 88.9\\
        \bottomrule
        
        \end{tabular}
    }
    \caption{\label{tab:abnormal} Performance for anomaly detection. }
\end{table}

\subsection{Effect of Randomization Strategy}
As the ablation study reveals, the frequency-aware randomization contributes the most to the defense. In this section, we analyze the impact of different hyperparameters and strategies adopted by the frequency-aware randomization approach, in inference and training respectively.

\subsubsection{Inference}
The frequency-aware randomization process is applied in inference to mitigate the adversarial effects. Substitution candidate selection and synonym set construction are two critical steps during this process, in which two hyperparameters ($r$ and $n_s$) and the frequency-aware strategy are examined.

\begin{figure}[h!]
	\centering
	\includegraphics[width=0.45\textwidth]{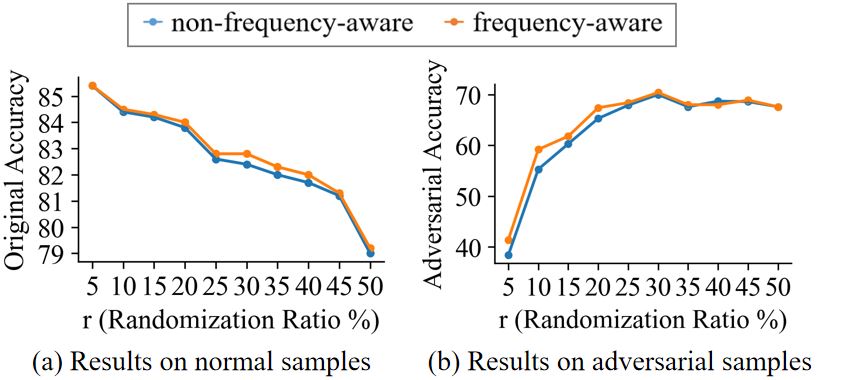}
	\caption{\label{fig:strategy_fig2} Effect of the substitution ratio $r$ and the frequency-aware strategy in substitution candidate selection during inference.
	}
\end{figure}

\paragraph{Selection of Substitution Candidates}
The influence of different strategies for substitution candidate selection in inference is studied in this section. The impact of two major factors are measured: 1) the substitution ratio $r$ and 2) whether to apply a frequency-aware strategy. In order to exclude the disturbance from other factors, we train BERT on the original training set and fix $n_s$ to $20$. Firstly, we alter the value of $r$ from $5\%$ to $50\%$, without applying the frequency-aware strategy. As illustrated by the blue lines in Figure \ref{fig:strategy_fig2}, as $r$ increases, the original accuracy decreases, while the adversarial accuracy increases and peaks when $r$ reaches $30\%$. Secondly, a frequency-aware strategy is added to the experiment, with $f_{thres} = 200$. As depicted by the yellow lines in Figure \ref{fig:strategy_fig2}, both original and adversarial accuracy, the general trends coincide with the non-frequency-aware scenario, but overall accuracy is improved to a higher level. The highest adversarial is obtained when  $r$ is set to $30\%$ using frequency-aware strategy.

\begin{figure}[h!]
	\centering
	\includegraphics[width=0.45\textwidth]{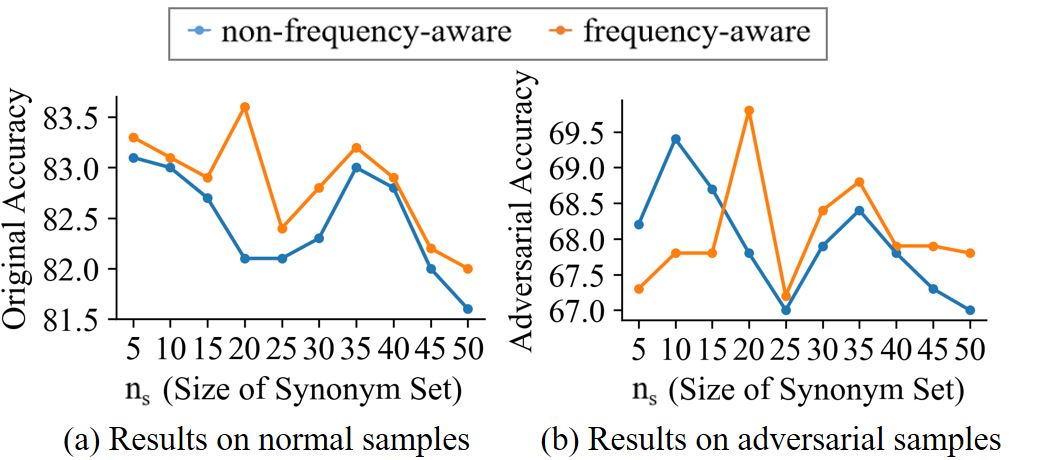}
	\caption{\label{fig:strategy_fig1} Effect of the size of synonym set $n_s$ and the frequency-aware strategy in construction of synonym set. 
	}
\end{figure}

\paragraph{Construction of Synonym Set}
The influence of different strategies for synonym set construction in inference is evaluated in this section. The impact of two major factors are measured: 1) the size of a single synonym set $n_s$ and 2) whether to apply a frequency-aware strategy. In order to exclude the disturbance from other factors, we train BERT on the original training set and fix $r$ to $30\%$ . Firstly, we alter the value of $n_s$ from $5$ to $50$, without applying the frequency-aware strategy. The resulted original and adversarial accuracy are illustrated by the blue lines in Figure \ref{fig:strategy_fig1}. Secondly, a frequency-aware strategy is added to the experiment, with $n_f = 50\%*n_s$. As depicted by the yellow lines in Figure \ref{fig:strategy_fig1}, the original accuracy and the adversarial accuracy both peaks when $n_s = 20$, and the overall accuracy is improved to a higher level compared to the non-frequency-aware scenario.

\subsubsection{Training}
\begin{figure}[h!]
	\centering
	\includegraphics[width=0.48\textwidth]{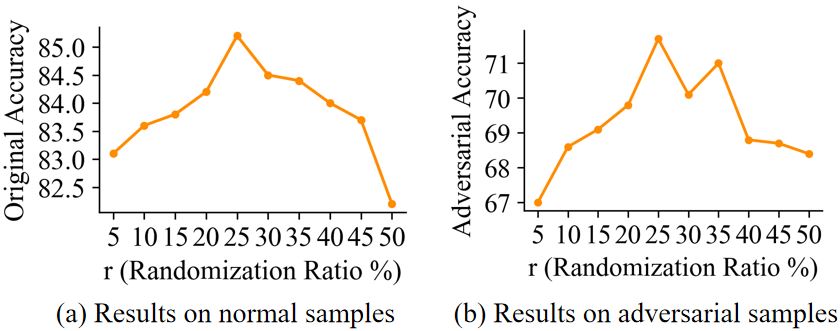}
	\caption{\label{fig:strategy_fig3} Effect of the size of synonym set $n_s$ and the frequency-aware strategy in construction of synonym set.
	}
\end{figure}

The frequency-aware randomization process is applied in training to augment the training data, and hereby enables the PrLM to better cope with randomized samples inference. Based on this purpose, the frequency-aware randomization process in training should resemble the one in inference as much as possible. Therefore, here we set an identical process for synonym set construction, i.e. $n_s=20$ and $n_f = 50\%*n_s$. However, for the substitution selection process, to avoid introducing excessive noise and maintain the accuracy for the PrLM, the most suitable substitution ratio $r$ might be different than the one in inference. Experiments are conducted to evaluate the influence of $r$ in training. We alter the value of $r$ from $5\%$ to $50\%$. In Figure \ref{fig:strategy_fig3}, we observe that $r = 25\%$ results in highest original and adversarial accuracy.

\section{Conclusion}
This paper proposes ADFAR, a novel framework which leverages the frequency-aware randomization and the anomaly detection to help PrLMs defend against adversarial word substitution. Empirical results show that ADFAR significantly outperforms those newly proposed defense methods over various tasks. Meanwhile, ADFAR achieves a remarkably higher inference speed and does not reduce the prediction accuracy for non-adversarial sentences, from which we keep the promise for this research purpose.

Comprehensive ablation study and analysis indicate that 1) Randomization is an effective method to defend against heuristic attack strategy. 2) Replacement of rare words with their more common alternative can help enhance the robustness of PrLMs. 3) Adversarial samples generated by current heuristic adversarial word substitution models can be easily distinguished by the proposed auxiliary anomaly detector. We hope this work could shed light on future studies on the robustness of PrLMs.

\bibliographystyle{acl_natbib}
\bibliography{anthology,acl2021}


\end{document}